\documentclass[conference]{IEEEtran}
\usepackage{cite}
\usepackage{makecell}
\usepackage{multirow}
\usepackage{amsmath,amssymb,amsfonts}
\usepackage{algorithmic}
\usepackage{graphicx}
\usepackage{siunitx}
\usepackage{numprint}
\usepackage{rotating} % Required for sideways figures
\usepackage{afterpage}
\usepackage[hidelinks]{hyperref}
\usepackage{textcomp}
\usepackage{xcolor}
\usepackage{booktabs}

\graphicspath{{figures/}}%it will look in the folder ``figures'' when calling ``\includegraphics{image.png}''

\def\BibTeX{{\rm B\kern-.05em{\sc i\kern-.025em b}\kern-.08em
    T\kern-.1667em\lower.7ex\hbox{E}\kern-.125emX}}
\begin{document}

% Cover page content
\begin{titlepage}
    \centering
    \vspace*{1.5in} % Add more space at the top for better presentation
    {\Huge \bfseries Unsupervised Clustering for Fault Analysis in High-Voltage Power Systems Using Voltage and Current Signals}\\[1.5cm]

    % Authors and Affiliations
    {\Large \bfseries Julian Oelhaf\textsuperscript{*}, Andreas Maier, Siming Bayer}\\
    \textit{Pattern Recognition Lab, Friedrich-Alexander-University Erlangen-Nuremberg}\\[0.5cm]

    {\Large \bfseries Georg Kordowich, Johann Jäger}\\
    \textit{Institute of Electrical Energy Systems, Friedrich-Alexander-University Erlangen-Nuremberg}\\[1.5cm]

    % Corresponding Author Information
    \textsuperscript{*}corresponding author: \texttt{julian.oelhaf@fau.de}\\[1cm]

    % Conference Details
    {\Large
    Presented at the \\
    \textbf{27th Annual Georgia Tech Fault and Disturbance Analysis Conference}\\
    Atlanta, Georgia\\
    May 5–6, 2025
    }
\end{titlepage}

\title{Unsupervised Clustering for Fault Analysis in High-Voltage Power Systems Using Voltage and Current Signals}

\author{
\IEEEauthorblockN{Julian Oelhaf\textsuperscript{1}\textsuperscript{*}, Georg Kordowich\textsuperscript{2}, Andreas Maier\textsuperscript{1}, Johann Jäger\textsuperscript{2}, Siming Bayer\textsuperscript{1}}
\IEEEauthorblockA{\textit{\textsuperscript{1}Pattern Recognition Lab, Friedrich-Alexander-University Erlangen-Nuremberg} \\
\textit{\textsuperscript{2}Institute of Electrical Energy Systems, Friedrich-Alexander-University Erlangen-Nuremberg} \\
Erlangen, Germany \\
\textsuperscript{*}corresponding author: julian.oelhaf@fau.de}
}
\maketitle

\begin{abstract}
The widespread use of sensors in modern power grids has led to the accumulation of large amounts of voltage and current waveform data, especially during fault events. However, the lack of labeled datasets poses a significant challenge for fault classification and analysis. This paper explores the application of unsupervised clustering techniques for fault diagnosis in high-voltage power systems. A dataset provided by the Réseau de Transport d'Électricité (RTE) is analyzed, with frequency domain features extracted using the Fast Fourier Transform (FFT). The K-Means algorithm is then applied to identify underlying patterns in the data, enabling automated fault categorization without the need for labeled training samples. The resulting clusters are evaluated in collaboration with power system experts to assess their alignment with real-world fault characteristics. The results demonstrate the potential of unsupervised learning for scalable and data-driven fault analysis by structuring large unlabeled waveform datasets and supporting targeted expert review.
\end{abstract}


\section{Introduction}\label{sec:intro}
The increasing deployment of sensors in modern power grids has led to the generation of vast amounts of data related to voltage and current waveforms, particularly during fault events. This growing availability of high-resolution measurements presents a unique opportunity for enhancing fault detection and diagnosis. However, a significant challenge lies in the fact that much of this data remains unlabeled, limiting its direct applicability in supervised learning-based fault classification. Extracting meaningful insights from these large-scale, unlabeled datasets is critical for ensuring grid stability and preventing cascading failures.

Accurate and timely fault diagnosis plays a pivotal role in power system operation, as undetected or misclassified faults can lead to severe system disturbances, increased downtime, and economic losses. Traditional fault classification methods typically rely on supervised machine learning models, which require extensive labeled datasets for training. However, generating such labeled datasets is both time-consuming and resource-intensive. This challenge is exacerbated in industrialized nations such as Germany, where a shortage of skilled professionals further hinders the manual labeling of power system faults. Consequently, there is a growing need for data-driven approaches that can extract insights from raw measurements without relying on expert-annotated datasets.

Unsupervised machine learning provides a promising alternative by enabling the automatic discovery of hidden patterns and structures within data. Clustering techniques, in particular, allow for the identification and categorization of fault events without predefined labels, making them well-suited for real-world power system applications. This paper applies unsupervised clustering to voltage and current signals to identify structure in high-voltage fault recordings without predefined labels.

To this end, a dataset provided by Réseau de Transport d'Électricité (RTE) is analyzed, with frequency-domain features extracted using the Fast Fourier Transform (FFT). The K-Means clustering algorithm is then employed to group fault events based on their underlying similarities. The resulting clusters are evaluated in collaboration with power system experts to assess their alignment with real-world fault characteristics. The findings highlight the potential of unsupervised learning in power system fault analysis by reducing reliance on labeled data and supporting scalable exploratory analysis of fault recordings.

The remainder of this paper is organized as follows: Section 2 describes the dataset and preprocessing steps. Section 3 provides an overview of fault types in transmission lines. Section 4 details the methodology, including feature extraction and clustering. Section 5 presents the clustering analysis, while Section 6 reports the results. The discussion interprets the findings and their implications, followed by the conclusion summarizing key insights and future directions.

\section{Dataset}\label{sec:dataset}

Understanding the characteristics of fault events in high-voltage power systems is crucial for developing reliable protection mechanisms. This study leverages a real-world dataset of voltage and current waveform recordings from fault events in the French transmission network. The dataset provides valuable insights into transient behavior and serves as the basis for applying unsupervised clustering techniques to fault analysis.
In this section, we describe the data acquisition process, key signal characteristics, and preprocessing steps.

\subsection{Data Source and Acquisition}
The dataset used in this study is provided by RTE\footnote{\href{https://www.rte-france.com/en/home}{https://www.rte-france.com/en/home}}, the French transmission system operator. It is available through the RTE Fault Recording Database (DFRDB)\footnote{\href{https://dfrdb.rte-france.com/}{https://dfrdb.rte-france.com/}} and comprises 12,053 voltage and current waveform recordings from France’s high-voltage transmission network~\cite{presvots_database_2024}. As the operator of more than \numprint{105000} kilometers of transmission lines and \numprint{50} interconnections with neighboring countries, RTE plays a key role in ensuring grid stability, making this dataset invaluable for improving fault analysis and protection strategies.

Each recording contains three-phase voltage and current signals. The signals are recorded at a high sampling rate of \(\qty{6400}{\hertz}\), enabling precise capture of transient voltage and current variations, which are essential for accurate fault analysis, classification, and diagnosis. Analyzing these signals is crucial for understanding fault dynamics and enhancing the reliability of protection mechanisms.

\subsection{Signal and Sample Characteristics}
The dataset consists of \numprint{12053} event recordings, each containing six time-synchronized signals: three phase voltages \((V_1, V_2, V_3)\) and three phase currents \((I_1, I_2, I_3)\). Each signal is recorded for \numprint{21000} timesteps, corresponding to  \(\qty{3.28125}{\second}\) of data.
The data is structured as a three-dimensional array with dimensions \((\numprint{12053}, \numprint{6}, \numprint{21000})\), where:
\begin{itemize}
    \item \numprint{12053} represents the number of recordings,
    \item \numprint{6} denotes the number of signals per recording (three voltage and three current signals),
    \item \numprint{21000} corresponds to the number of timesteps per signal.
\end{itemize}

The nominal system frequency is \(\qty{50}{\hertz}\), and the nominal voltage level is \(\qty{90}{\kilo\volt}\), reflecting the operating conditions of the French high-voltage transmission network. The dataset was preselected based on waveform characteristics to include only significant recorded events. This selection was performed by the dataset providers, and we have no insight into their specific criteria or process.

The signals are recorded at a sampling frequency of \(\qty{6400}{\hertz}\), providing high temporal resolution for capturing rapid voltage and current transients during faults. They are quantized using 16-bit encoding, with amplitude values ranging from \numprint{-32767} to \numprint{32767}, ensuring precise representation of both minor and severe disturbances.
The quantization step sizes are \(\qty{18.310}{\volt}\) for voltage signals and \(\qty{4.314}{\ampere}\) for current signals, ensuring accurate representation of signal variations critical for fault analysis. 

Notably, the dataset lacks explicit labels or background information on fault types, motivating the use of an unsupervised approach for fault analysis and classification. The only available information comes from the RTE Fault Recording Database (DFRDB)~\cite{presvots_database_2024}.

\subsection{Data Preprocessing}

Prior to analysis, the voltage and current waveforms undergo preprocessing to enhance interpretability and improve the effectiveness of clustering techniques. This process consists of two primary steps: normalization and decomposition-based anomaly detection.

To ensure consistency across all recordings, each signal is independently normalized to a fixed range. For a given raw discrete-time signal \( X[n] \), its normalized counterpart \( \tilde{X}[n] \) is computed as:

\begin{equation}
    \tilde{X}[n] = 2 \frac{X[n] - X_{\min}}{X_{\max} - X_{\min}} - 1,
\end{equation}

where \( X_{\min} \) and \( X_{\max} \) represent the minimum and maximum values of the signal, respectively. This transformation scales the signal to the range \([-1,1]\), ensuring numerical stability and enabling comparison across different recordings.

Each signal is then decomposed into three components: the trend \( T[n] \), which represents long-term variations; the seasonal component \( S[n] \), capturing periodic fluctuations related to the nominal power system frequency; and the residual \( e[n] \), which contains transient deviations and noise. The decomposition follows an additive model:

\begin{equation}
    Y[n] = T[n] + S[n] + e[n],
\end{equation}

where \( Y[n] \) is the original sampled signal. The trend component \( T[n] \) is estimated using a moving average filter, which smooths long-term variations.
After removing the trend, the seasonal component \( S[n] \) is computed as the mean of the de-trended series over one full cycle of the power system’s fundamental frequency.
Given the system’s nominal frequency of \(\qty{50}{\hertz}\) and a sampling rate of \(\qty{6400}{\hertz}\), the decomposition period is \(128\) time steps.
Seasonal decomposition is performed using the \texttt{seasonal\_decompose} class provided in the \texttt{statsmodels} library~\cite{seabold2010statsmodels}. In addition to the decomposition, zero-voltage \( Z_V[n] \) and zero-current \( Z_I[n] \) indicators are computed for each sample. 

Finally, anomaly detection is applied to the voltage signal, specifically targeting the trend \( T[n] \), residual \( e[n] \), and zero-voltage \( Z_V[n] \) components, in order to identify time intervals with significant deviations from normal operation. Thresholds based on statistical measures are defined for each component to highlight regions where abnormal transients or zero-voltage events occur. This enables the isolation of periods where the signal deviates from expected behavior, indicating potential faults or disturbances in the system. While decomposition and anomaly detection are not used in the clustering process, the detected anomalies support the manual visual analysis by highlighting regions of interest.

This preprocessing pipeline structures the dataset to emphasize significant variations while filtering out noise and redundant information, ensuring that clustering techniques can effectively distinguish between different fault behaviors based on intrinsic signal characteristics.

\section{Overview of Fault Types in Transmission Lines}

Faults in transmission lines are undesirable short-circuit conditions that can lead to significant damage in power systems. These faults can be classified based on their electrical characteristics and underlying causes. A thorough understanding of fault types is essential for the development of effective diagnostic and prognostic techniques in high-voltage and medium-voltage power networks \cite{bindi_comprehensive_2023}.

\subsubsection*{Overvoltages}

Overvoltages occur when the system voltage exceeds the nominal design value. They can be either external or internal in nature. External overvoltages are primarily caused by lightning strikes, which can lead to insulation failure and catastrophic short-circuits. Internal overvoltages are caused by switching operations, load variations, or resonance phenomena, and they can lead to insulation breakdown within the system.

\subsubsection*{Overcurrents}

Overcurrents occur when the electrical current surpasses the system's designed limits. These can be transient, permanent, or fault-induced in nature. Transient overcurrents, such as those arising from motor startups or transformer energization, are typically short-lived and do not require circuit breaker intervention. Permanent overcurrents, on the other hand, occur when sustained overloads lead to conductor heating and potential long-term damage. Fault-induced overcurrents result from faults like insulation failure or accidental contact, generating high short-circuit currents that can cause extensive damage to system components.

\subsubsection*{Short-Circuit Faults}

Short-circuit faults are among the most critical types of faults in transmission systems. They involve unintended low-impedance connections between conductors or between a conductor and the ground. Short-circuit faults can be categorized into four main types. The most common is the single-line-to-ground (SLG) fault, which occurs when one phase conductor comes into contact with the ground. Line-to-line (LL) faults involve a short-circuit between two phase conductors, typically due to insulation failure or mechanical damage. Double-line-to-ground (DLG) faults, where two phase conductors simultaneously make contact with the ground, are more severe and result in high fault currents. The most catastrophic fault is the three-phase (LLL) fault, which involves all three phases and typically results from equipment failure or severe external disturbances.

\subsubsection*{Open Circuit Faults}

Open circuit faults occur when the conducting path is broken, often due to conductor failure or cable joint issues. These faults can lead to unbalanced currents within the system, which can cause overheating of rotating machinery and other operational disturbances. 

\subsubsection*{Power Quality Disturbances}

In addition to faults, power transmission systems are subject to various power quality disturbances that can affect system reliability. These disturbances include voltage fluctuations, which arise from changing load conditions or renewable energy integration, and flicker, which is caused by short-term voltage variations from industrial loads or renewable generation. Harmonic distortion, resulting from nonlinear loads and power electronic converters, can also cause significant waveform distortions, interfering with grid operation and equipment performance \cite{bindi_comprehensive_2023}.

Effective detection, diagnosis, and protection against faults and power quality disturbances are essential for maintaining the stability and resilience of power transmission systems. The integration of advanced fault diagnosis techniques, including artificial intelligence-based approaches, plays a vital role in ensuring operational efficiency and grid stability.

\section{Methodology}\label{sec:methodology}

The proposed approach consists of three main stages: (i) feature extraction in the frequency domain using the Fourier Transform, (ii) dimensionality reduction through two distinct techniques, and (iii) clustering of the reduced feature representations using the K-Means algorithm.

\subsection{Feature Extraction}

To transform raw time-series data for machine learning, we apply the \textit{Fast Fourier Transform (FFT)} to extract frequency-domain features. The FFT converts a signal \( x(n) \) from the time domain into its frequency components, capturing periodic patterns that may indicate faults. The FFT is computed as:

\begin{equation}
    X(f) = \sum_{n=0}^{N-1} x(n) e^{-j 2\pi f n / N}
\end{equation}

We retain only the positive frequencies (first \( N/2 \) coefficients), as the FFT output is symmetric for real-valued signals. FFT is chosen for its efficiency in revealing spectral patterns associated with power system disturbances, such as impedance changes, arcing, and harmonics from faults.

\subsubsection*{Feature Selection and Normalization}

To standardize the features, FFT magnitudes are normalized by the maximum magnitude:

\begin{equation}
    X_{\text{norm}}(f) = \frac{|X(f)|}{\max_{f'} |X(f')|}
\end{equation}

Normalization reduces sensitivity to absolute amplitude differences and emphasizes relative spectral characteristics. However, it may suppress information on absolute magnitude changes.

\subsubsection*{Domain-Specific Considerations}

FFT aligns with established power system research, where faults exhibit distinct frequency-domain patterns:

\begin{itemize}
    \item \textbf{Short circuits} show high-frequency transients due to sudden current flow changes.
    \item \textbf{Open-circuit faults} produce dominant low-frequency components from power loss.
    \item \textbf{Unbalanced faults} cause asymmetries captured as \textit{negative sequence components}.
\end{itemize}

These insights make the extracted features meaningful representations of fault behaviors in electrical networks.

\subsection{Feature Reduction}
\subsubsection*{Principal Component Analysis (PCA)}

PCA is a linear technique that transforms the original feature space into a new coordinate system defined by the principal components, which are directions of maximum variance in the data \cite{greenacre_principal_2022}. It reduces the dimensionality by projecting the original feature matrix \( \mathbf{X} \in \mathbb{R}^{N \times D} \) onto a matrix of principal components \( \mathbf{P} \in \mathbb{R}^{D \times K} \), where \( K \ll D \):

\begin{equation}
    \mathbf{X}_{\text{reduced}} = \mathbf{X} \mathbf{P}
\end{equation}

PCA retains the most significant components. In this study we retain \(3\) components of the \numprint{63000}-dimensional FFT feature vector. These account for \(32.7\%\), \(12.0\%\) and \(8.1\%\) of the variance respectively, \(52.7\%\) in total. The component count was fixed a priori at three for the subsequent dimensionality-reduction and clustering pipeline, rather than chosen to reach a variance target, so roughly half of the spectral variance is not represented in the clustering space. The implementation uses the \texttt{PCA} class from \texttt{scikit-learn}~\cite{pedregosa_scikit-learn_2011}.

\subsubsection*{t-Distributed Stochastic Neighbor Embedding (t-SNE)}

While PCA handles linear reductions, t-SNE addresses non-linear relationships in high-dimensional data. t-SNE visualizes data by converting pairwise distances into conditional probabilities and minimizing the Kullback-Leibler (KL) divergence between high- and low-dimensional probability distributions \cite{maaten_visualizing_2008}. t-SNE is implemented in the \texttt{TSNE} class of \texttt{scikit-learn}~\cite{pedregosa_scikit-learn_2011}. t-SNE first defines conditional similarities in the high-dimensional space as

\begin{equation}
    p_{j|i} =
    \frac{\exp\left(-\| \mathbf{x}_i-\mathbf{x}_j \|^2 / 2\sigma_i^2\right)}
    {\sum_{k \neq i}\exp\left(-\| \mathbf{x}_i-\mathbf{x}_k \|^2 / 2\sigma_i^2\right)},
\end{equation}

with \(p_{i|i}=0\). These conditional probabilities are symmetrized as

\begin{equation}
    p_{ij} = \frac{p_{j|i}+p_{i|j}}{2N}.
    \label{eq:p_ij}
\end{equation}

t-SNE then maps the samples to a lower-dimensional space \( \mathbf{Y} \in \mathbb{R}^{N \times 2} \) or \( \mathbb{R}^{N \times 3} \) and defines similarities using a normalized Student-\(t\) kernel:

\begin{equation}
    q_{ij} =
    \frac{\left(1+\| \mathbf{y}_i-\mathbf{y}_j \|^2\right)^{-1}}
    {\sum_{k \neq l}\left(1+\| \mathbf{y}_k-\mathbf{y}_l \|^2\right)^{-1}},
    \label{eq:q_ij}
\end{equation}

with \(q_{ii}=0\).

\subsubsection*{Application in this Study}

PCA is initially applied to reduce the dimensionality of the FFT features to three components. Following PCA, t-SNE is used to further reduce the data. In this study, t-SNE reduces the PCA representation to three dimensions for clustering and analysis; the two-dimensional views in Fig.~\ref{fig:cluster_visualization_2d} are projections of that same three-dimensional representation, used for visualization only. This combination aids in the identification and interpretation of fault-related structures within the data.

\subsection{Clustering Algorithm}

Clustering is essential for identifying patterns in high-dimensional data. K-Means Clustering, known for its simplicity, efficiency, and interpretability, is widely used for partitioning fault patterns in high-dimensional feature spaces. First proposed by Stuart Lloyd in 1957 for pulse-code modulation, the term "K-Means" was introduced by James MacQueen in 1967. The algorithm is also called the Lloyd–Forgy method, as Edward W. Forgy independently derived it in 1965~\cite{lloyd_least_1982}.

The K-Means algorithm aims to partition a dataset \( \mathbf{X} \in \mathbb{R}^{N \times D} \) into \( K \) distinct clusters. The objective is to minimize the within-cluster sum of squared distances between data points and the centroids of their respective clusters. Given a set of data points, K-Means performs the following steps iteratively:

\begin{itemize}
    \item \textbf{Initialization}: Randomly select \( K \) initial cluster centroids.
    \item \textbf{Assignment Step}: Assign each data point to the closest centroid based on a distance metric (typically Euclidean distance).
    \item \textbf{Update Step}: Recompute the centroids by calculating the mean of all points assigned to each centroid.
    \item \textbf{Convergence}: Repeat the assignment and update steps until convergence, typically when the centroids no longer change or the maximum number of iterations is reached.
\end{itemize}

Mathematically, the objective function that K-Means seeks to minimize is given by:

\begin{equation}
    J = \sum_{k=1}^{K} \sum_{i=1}^{N_k} \| \mathbf{x_i} - \mathbf{\mu_k} \|^2
    \label{eq:kmeans_objective}
\end{equation}

where \( \mathbf{x_i} \) represents the \( i \)-th data point, \( \mathbf{\mu_k} \) is the centroid of the \( k \)-th cluster, and \( N_k \) is the number of data points assigned to cluster \( k \). The term \( \| \mathbf{x_i} - \mathbf{\mu_k} \|^2 \) represents the squared Euclidean distance between the data point and the centroid.

\paragraph*{Implementation and Parameters}

The K-Means algorithm is implemented using the \texttt{KMeans} class in the \texttt{scikit-learn} library \cite{pedregosa_scikit-learn_2011}. The algorithm requires the specification of the number of clusters \( K \), which is often determined using methods such as the elbow method or silhouette analysis. The elbow method involves plotting the sum of squared distances (within-cluster inertia) for different values of \( K \) and selecting the value at which the rate of decrease slows down. The silhouette analysis measures how similar a point is to its own cluster compared to other clusters, providing a quantitative way to assess the quality of clustering.

\paragraph*{Applications and Limitations}

K-Means is widely used due to its computational efficiency, especially for large datasets. It is particularly well-suited for scenarios where the clusters are approximately spherical and have roughly equal sizes. This makes K-Means effective for partitioning fault patterns that exhibit well-separated, homogeneous characteristics in feature space.

However, K-Means has several limitations. One major limitation is its sensitivity to the initial placement of centroids. Poor initialization can lead to suboptimal clustering results. Additionally, K-Means assumes that clusters are spherical and equally sized, which may not hold true for all datasets. To mitigate these limitations, variations of K-Means, such as K-Means++ for centroid initialization, have been proposed.

\paragraph*{Usage in this Study}

In this study, K-Means is applied to the extracted features to identify fault patterns in the high-dimensional feature space. After dimensionality reduction using PCA and t-SNE, K-Means is employed to partition the data into distinct clusters that may capture different fault or event patterns. The resulting clusters are analyzed to identify key patterns and relationships between the fault occurrences, which can then inform further diagnostic and predictive modeling.

\section{Clustering Analysis}
In this study, we adopt a two-stage approach to examine the clustering of fault signals. First, we quantitatively assess the clustering output using metrics such as cluster separability and sample distribution. This analysis characterizes the resulting groups quantitatively. Second, we perform an expert-driven validation where a power system specialist reviews representative samples from each cluster. This step enables us to interpret cluster compositions, confirm correspondence with actual fault types, and identify any overlaps among fault categories.

\begin{figure*}[!htbp]
	\centering
	\includegraphics[trim={0 0 0 0},clip,width=0.98\linewidth]{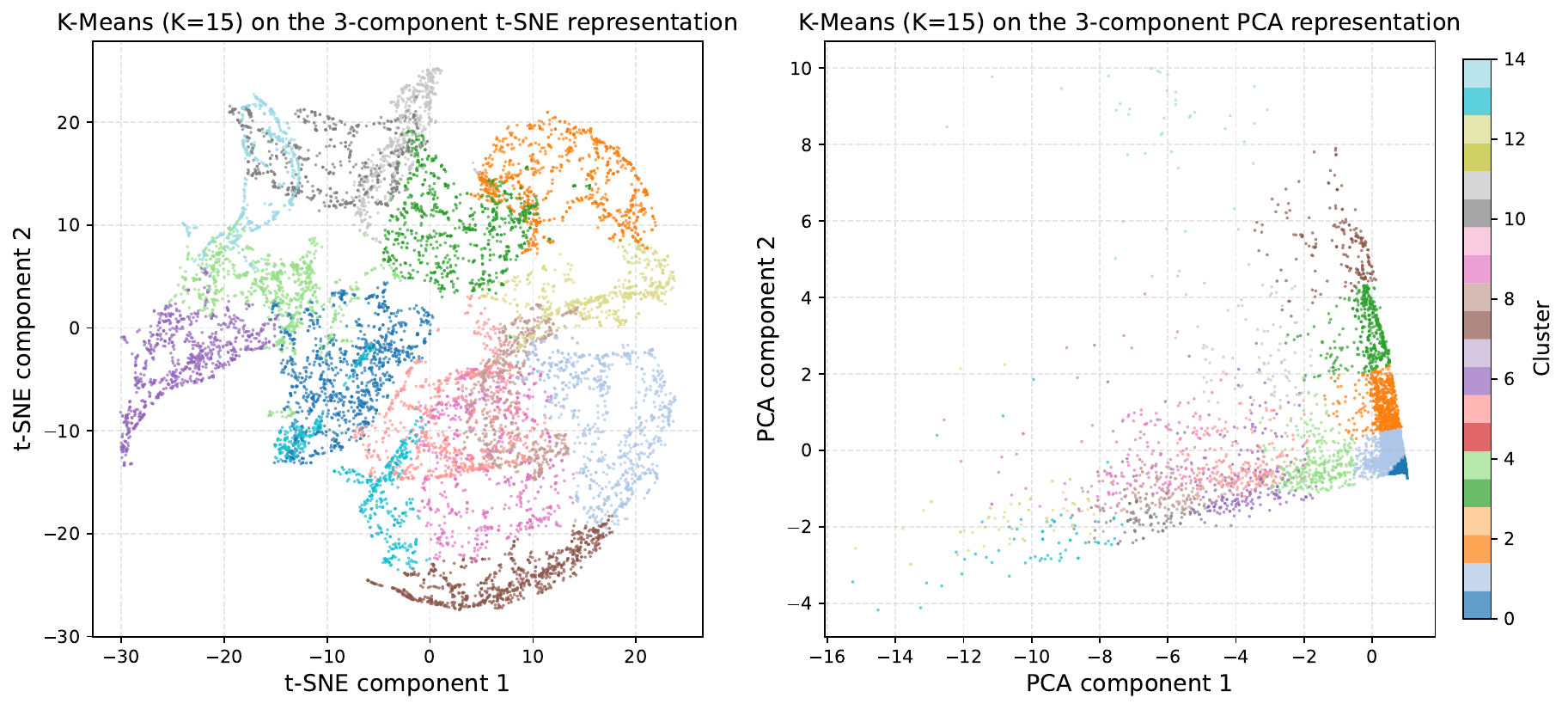}
    \caption{Two-dimensional view of the clustering results. Each point is one of the \numprint{12053} recordings, colored by its K-Means cluster. Left: the 3-component t-SNE representation; right: the 3-component PCA representation. In both panels the clusters are the ones used throughout this paper, obtained by K-Means in the full 3-component space; only the \emph{view} is two-dimensional, taking the first two coordinates. No separate clustering was performed for this figure.}  
    \label{fig:cluster_visualization_2d}
\end{figure*}

\subsection{Cluster Visualization}

Figure~\ref{fig:cluster_visualization_2d} shows two-dimensional views of the three-component t-SNE and PCA representations, with samples colored according to the K-Means assignments obtained in the respective full three-component spaces. This visualization provides an intuitive assessment of cluster separability, helping to identify both well-separated clusters and overlapping regions that may indicate transitional states or feature ambiguities. Whether the separated regions correspond to distinct fault or event types is not established by the visualization; it is assessed below through human validation. For further evaluation, we consider two key aspects:

\textbf{Cluster Separability:} This assesses the spatial distinctiveness of the clusters. Well-separated clusters indicate distinct regions of the feature space; whether these regions correspond to distinct fault or event types requires expert validation. Overlap may indicate similarities or transitional conditions between fault states.

\textbf{Cluster Size Distribution:} This examines the number of samples within each cluster. Larger clusters typically correspond to more common fault patterns, while smaller clusters may represent rare or anomalous events.

The two-dimensional views in Fig.~\ref{fig:cluster_visualization_2d} offer valuable insights into the clustering results. Combined with quantitative metrics, these visual tools allow for a comprehensive evaluation of clustering quality.

\subsection{Manual Expert Labeling}

To further validate the clustering results, a power system expert manually examines a subset of signals from each cluster. During this process, the expert assigns fault labels based on established fault characteristics, comparing the unsupervised clusters with real-world fault types such as three-phase short circuits (3PH-SC), line-to-ground (LG) faults, and line-to-line (LL) faults.

For this evaluation, randomly selected samples from each cluster are reviewed. The expert assesses cluster coherence by visualizing the voltage and current signals over multiple time windows, enabling a detailed analysis of transient phenomena, waveform distortions, and phase interactions. Each labeled sample is annotated with the following details:
\begin{itemize}
    \item \textbf{Fault Class:} A broad category (e.g., short circuit, open circuit, transient disturbance).
    \item \textbf{Fault Type:} A specific classification (e.g., single-phase-to-ground short circuit, three-phase fault).
    \item \textbf{Phase:} The impacted phase(s).
    \item \textbf{Comment:} Additional observations or clarifications regarding the fault characteristics.
\end{itemize}

\begin{table}[!tb]
    \centering
    \caption{Example of Labels Assigned by the Expert.}
    \resizebox{0.98\linewidth}{!}{
    \begin{tabular}{r|lll}
        \toprule
         \textbf{Sample ID} & \textbf{Fault Class} & \textbf{Fault Type} & \textbf{Phase} \\
        \midrule
        27 & Normal & Normal & N/A \\
        2654 & Other & Off - No Switch & N/A \\
        3335 & Short-circuit  & 1-P-SC & C \\
        5125 & Transients & Transients & N/A \\
        8578 & Switching & Switch On & N/A \\
        11491 & Short-circuit & 1-P-SC & B \\
        \bottomrule
    \end{tabular}
    }
    \label{tab:example_labels}
\end{table}

For example, a sample might be labeled as:
\textit{Fault Class: Short Circuit, Fault Type: 1-P-SC, Phase: A}, accompanied by comments describing waveform irregularities or recovery patterns. An excerpt of the labels assigned by the expert is shown in Table~\ref{tab:example_labels}.

Additionally, the expert is provided with an interactive feature that allows them to \textit{scroll} through the signal data. This enables seamless switching between detailed, zoomed-in views and comprehensive full-sequence plots, ensuring that localized anomalies are evaluated within the broader context of the fault event. This multi-scale assessment enhances labeling accuracy and provides a valuable reference for improving fault classification models in high-voltage power systems. Figure~\ref{fig:voltage_current_analysis} in the Appendix illustrates an example highlighting transient effects and the subsequent post-fault stabilization.

\section{Clustering Results}  

In this work, clustering is employed to identify distinct fault and event patterns within high-voltage power systems. The number of clusters is set to \(15\), reflecting the diversity of known fault types and system disturbances; it was chosen a priori rather than by an elbow or silhouette criterion. All reported results originate from a single clustering run (\(K=15\), three components, fixed random seed), fitted on the archived reduced feature representations; the cluster sizes, the evaluation metrics, the heatmaps and the visualization in Fig.~\ref{fig:cluster_visualization_2d} are therefore mutually consistent. The choice of \(K=15\) balances interpretability and granularity, enabling meaningful distinctions between different event categories.

\subsection{Overview of Clustering Performance}  

Tables~\ref{tab:number_samples_per_cluster_pca} and~\ref{tab:number_samples_per_cluster_tsne} present the distribution of samples across clusters, obtained using K-Means with PCA and t-SNE for dimensionality reduction, respectively. The PCA-based approach results in highly imbalanced cluster sizes, with the largest cluster containing \numprint{5288} samples (\(43.9\%\)) and the smallest only \(38\) (\(0.3\%\)). The standard deviation in cluster sizes is \numprint{1485} (\(12.32\%\) of the total dataset), indicating that a few large clusters dominate the partitioning.

Conversely, the t-SNE-based clustering produces a more uniform distribution, with cluster sizes ranging from \(574\) (\(4.8\%\)) to \numprint{1098} (\(9.1\%\)) samples, and a significantly lower standard deviation of \numprint{137} (\(1.14\%\)). These differences underscore the impact of dimensionality reduction: while PCA preserves global variance leading to imbalanced clusters, t-SNE emphasizes local structure, resulting in a more balanced clustering outcome.

% Generated by fault_cluster.reports.fda2025 -- do not edit by hand.
% Canonical run: 2026-09-09_fda2025_v2_8a2c1d279141
\begin{table*}[!tb]
    \centering
    \caption{Number of Samples per Cluster Using K-Means PCA Labels (\(N = 12,053\))}
    \begin{tabular}{cccccccccccccccc}
        \toprule
         & \textbf{0} & \textbf{1} & \textbf{2} & \textbf{3} & \textbf{4} & \textbf{5} & \textbf{6} & \textbf{7} & \textbf{8} & \textbf{9} & \textbf{10} & \textbf{11} & \textbf{12} & \textbf{13} & \textbf{14} \\
        \midrule
        \textbf{Count} & 5288 & 3665 & 882 & 467 & 367 & 255 & 200 & 199 & 171 & 170 & 125 & 95 & 68 & 63 & 38 \\
        \textbf{\%} & 43.9\% & 30.4\% & 7.3\% & 3.9\% & 3.0\% & 2.1\% & 1.7\% & 1.7\% & 1.4\% & 1.4\% & 1.0\% & 0.8\% & 0.6\% & 0.5\% & 0.3\% \\
        \bottomrule
    \end{tabular}
    \label{tab:number_samples_per_cluster_pca}
\end{table*}

% Generated by fault_cluster.reports.fda2025 -- do not edit by hand.
% Canonical run: 2026-09-09_fda2025_v2_8a2c1d279141
\begin{table*}[!tb]
    \centering
    \caption{Number of Samples per Cluster Using K-Means t-SNE Labels (\(N = 12,053\))}
    \begin{tabular}{cccccccccccccccc}
        \toprule
         & \textbf{0} & \textbf{1} & \textbf{2} & \textbf{3} & \textbf{4} & \textbf{5} & \textbf{6} & \textbf{7} & \textbf{8} & \textbf{9} & \textbf{10} & \textbf{11} & \textbf{12} & \textbf{13} & \textbf{14} \\
        \midrule
        \textbf{Count} & 1098 & 947 & 937 & 920 & 874 & 861 & 838 & 812 & 771 & 766 & 694 & 668 & 654 & 639 & 574 \\
        \textbf{\%} & 9.1\% & 7.9\% & 7.8\% & 7.6\% & 7.3\% & 7.1\% & 7.0\% & 6.7\% & 6.4\% & 6.4\% & 5.8\% & 5.5\% & 5.4\% & 5.3\% & 4.8\% \\
        \bottomrule
    \end{tabular}
    \label{tab:number_samples_per_cluster_tsne}
\end{table*}

\begin{figure}[!tbp]
    \centering
    \includegraphics[width=0.98\linewidth]{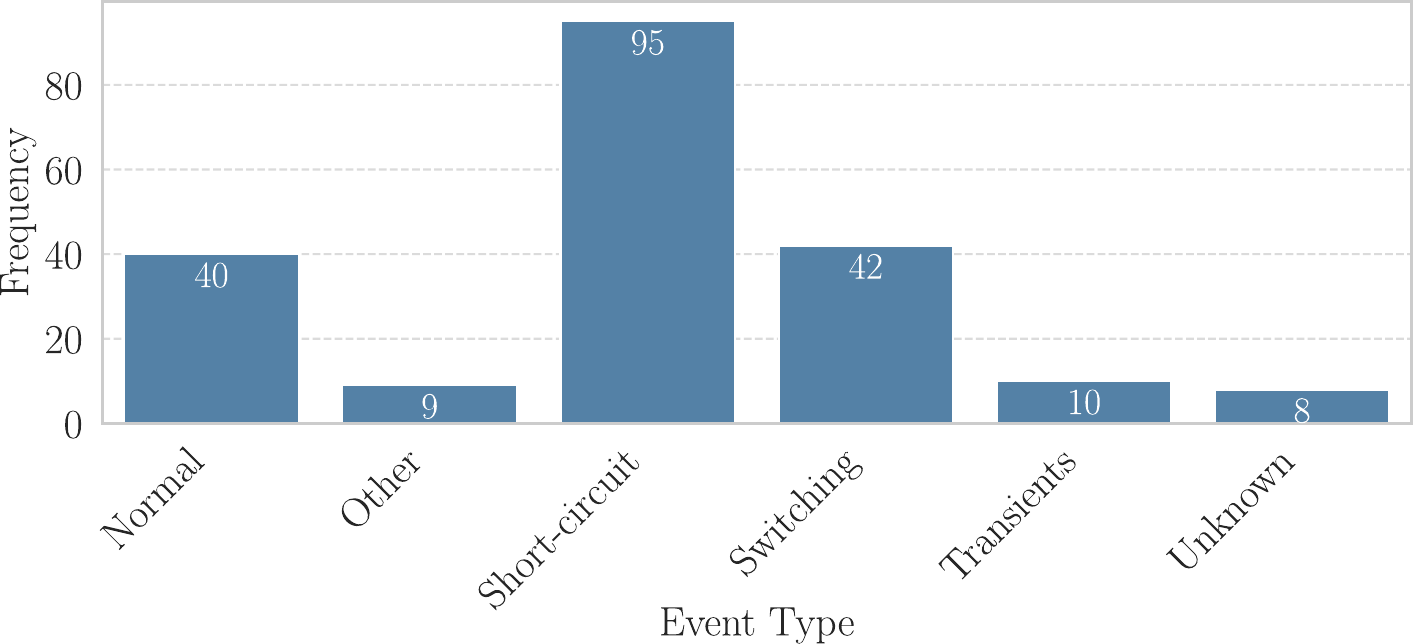}
    \caption{Distribution of the event types in the manually labeled subset (N = 204).}
    \label{fig:event_type_histogram}
\end{figure}

Figure~\ref{fig:event_type_histogram} shows the distribution of event types within the manually labeled subset (N = \(204\)), providing context for the fault categorization.

% Generated by fault_cluster.reports.fda2025 -- do not edit by hand.
% Canonical run: 2026-09-09_fda2025_v2_8a2c1d279141
% All three metrics are size-weighted over clusters.
\begin{table}[!tb]
    \caption{Global Clustering Evaluation Metrics (\(N=204\))}
    \label{tab:global_clustering_metrics}
    \centering
    \begin{tabular}{lrrr}
        \toprule
        \textbf{Clustering Method} & \makecell{\textbf{Purity} \\ ($\uparrow$)} & \makecell{\textbf{Entropy} \\ ($\downarrow$)} & \makecell{\textbf{Silhouette} \\ ($\uparrow$)} \\
        \midrule
        K-Means PCA & 0.608 & 1.336 & 0.487 \\
        K-Means t-SNE & 0.721 & 1.073 & 0.315 \\
        \bottomrule
    \end{tabular}
\end{table}

\subsection{Quantitative Evaluation of Clustering}

We evaluate the clustering quality using several standard metrics~\cite{manning_introduction_2008, cover_elements_2006, vinh_information_2010, hubert_comparing_1985, rousseeuw_silhouettes_1987}:

\textbf{Purity} measures how well each cluster consists of a single fault type:
\begin{equation}
\text{Purity} = \frac{1}{N} \sum_{k} \max_{j} \left| C_{k} \cap T_{j} \right|,
\end{equation}
where \(C_k\) is the set of samples in cluster \(k\), \(T_j\) is the set of samples of fault type \(j\), and \(N\) is the total number of samples. Higher values (closer to 1) indicate better purity.

\textbf{Entropy} quantifies uncertainty within a cluster:
\begin{equation}
H(C_k) = -\sum_{j} p_j \log_2 p_j,
\end{equation}
where \(p_j\) is the proportion of fault type \(j\) in cluster \(C_k\). Lower values are better, with 0 indicating a homogeneous cluster.

\textbf{Silhouette Score} measures the compactness and separation of clusters, ranging from -1 to 1. Higher values indicate well-defined, distinct clusters, while values closer to 0 suggest overlapping clusters, and negative values indicate poor clustering.

These metrics were computed for both PCA and t-SNE clustering methods, with results summarized in Tab.~\ref{tab:global_clustering_metrics}. Detailed results for each cluster are available in Appendix Tab.~\ref{tab:per_cluster_clustering_metrics}.

\subsection{Analysis of Manually Labeled Clustering Results}
Before clustering, \(100\) randomly selected samples are manually labeled by an expert to gain an initial understanding of the signals. Additionally, after clustering, \(7\) samples from each of the \(15\) clusters are randomly selected (one sample was excluded as it was unusable). This results in \(N=204\) manually labeled samples, enabling further analysis of the clusters. The fault-class analysis uses the \(200\) reviewed samples with an available fault-class annotation.

\begin{figure}[!tbp]
    \centering
    \includegraphics[width=0.7\linewidth]{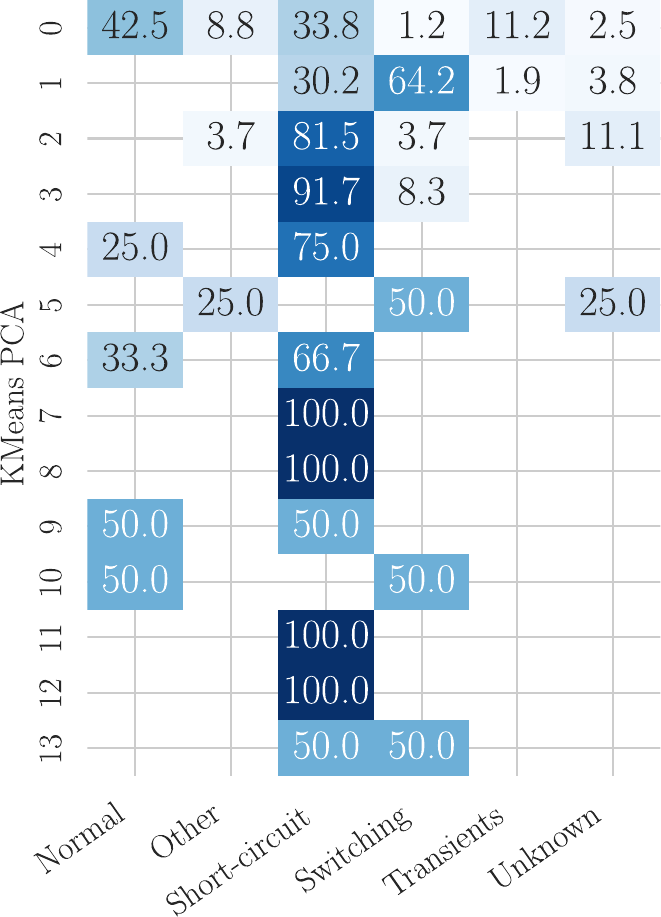}
    \caption{Percentage distribution of event types per K-Means PCA cluster in the manually labeled subset (N = 204). Each cluster sums to \(100\%\).}
    \label{fig:event_type_vs_kmeans_pca_heatmap_percentage}
\end{figure}

\begin{figure}[!tbp]
    \centering
    \includegraphics[width=0.7\linewidth]{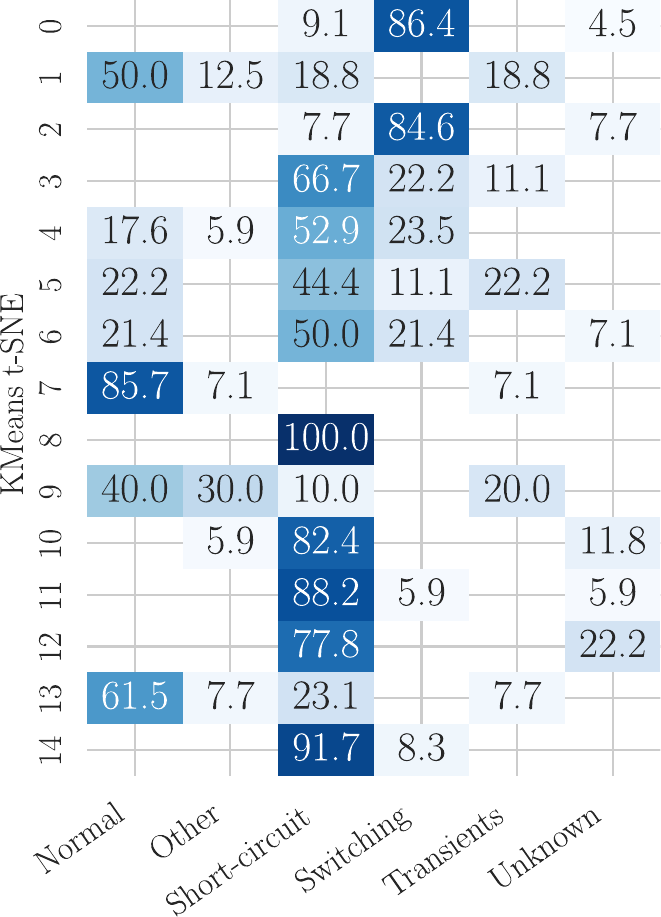}
    \caption{Percentage distribution of event types per K-Means t-SNE cluster in the manually labeled subset (\(N=204\)). Each cluster sums to \(100\%\).}
    \label{fig:event_type_vs_kmeans_tsne_heatmap_percentage}
\end{figure}

Fig.~\ref{fig:event_type_vs_kmeans_pca_heatmap_percentage} and Fig.~\ref{fig:event_type_vs_kmeans_tsne_heatmap_percentage} present heatmaps showing the percentage distribution of event types within each cluster for the PCA-based and t-SNE-based approaches, respectively. In each case, the percentages per cluster sum to \(100\%\), providing a visual representation of how faults are distributed across clusters.

The results highlight the impact of the chosen dimensionality reduction technique on clustering performance and provide insights into the internal consistency and separation of the identified clusters.

\begin{figure}[!tbp]
    \centering
    \includegraphics[width=0.98\linewidth]{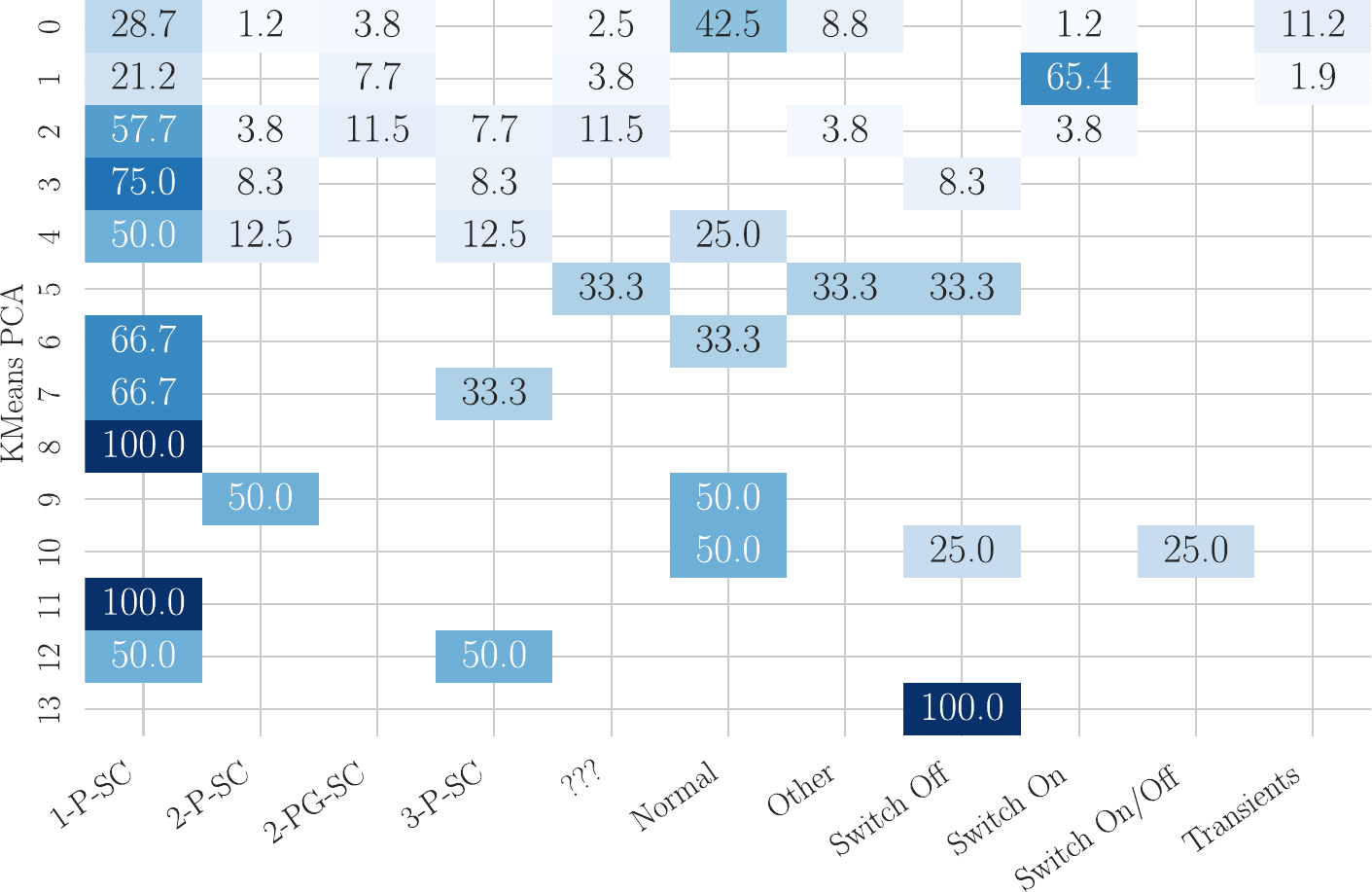}
    \caption{Percentage distribution of fault classes per K-Means PCA cluster in the manually labeled subset (\(N=200\)). Each cluster sums to \(100\%\).}
    \label{fig:fault_class_vs_kmeans_pca_heatmap_percentage}
\end{figure}

\begin{figure}[!tbp]
    \centering
    \includegraphics[width=0.98\linewidth]{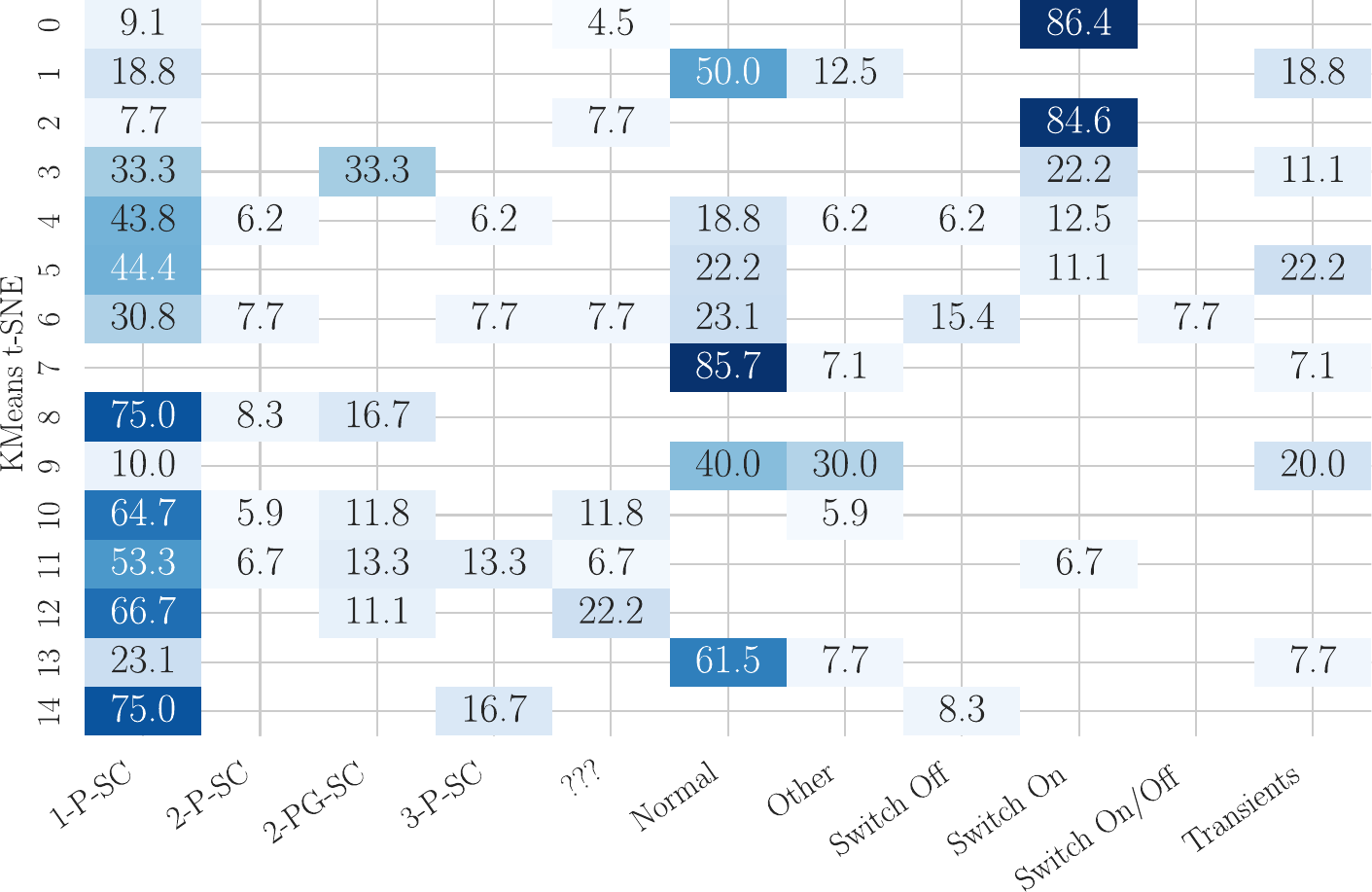}
    \caption{Percentage distribution of fault classes per K-Means t-SNE cluster in the manually labeled subset (\(N=200\)). Each cluster sums to \(100\%\).}
    \label{fig:fault_class_vs_kmeans_tsne_heatmap_percentage}
\end{figure}

% Generated by fault_cluster.reports.fda2025 -- do not edit by hand.
% Canonical run: 2026-09-09_fda2025_v2_8a2c1d279141
% Weighted and unweighted conventions occupy separate rows and are never mixed.
% Each cell is mean +- dispersion over clusters, using the same weighting as the
% mean it accompanies (see fault_cluster.analysis.metrics).
\begin{table*}[!tb]
    \centering
    \caption{Clustering metrics under both aggregation conventions. Values are reported as mean $\pm$ standard deviation over clusters, each dispersion using the same weighting as the mean it accompanies.}
    \label{tab:clustering_metrics}
    \begin{tabular}{llrrr}
        \toprule
        \textbf{Clustering Method} & \textbf{Aggregation} & \textbf{Purity} & \textbf{Entropy} & \textbf{Silhouette} \\
        \midrule
        K-Means PCA & size-weighted & \(0.608 \pm 0.182\) & \(1.336 \pm 0.555\) & \(0.487 \pm 0.212\) \\
        K-Means t-SNE & size-weighted & \(0.721 \pm 0.184\) & \(1.073 \pm 0.557\) & \(0.315 \pm 0.085\) \\
        \midrule
        K-Means PCA & unweighted cluster mean & \(0.730 \pm 0.216\) & \(0.767 \pm 0.585\) & \(0.371 \pm 0.201\) \\
        K-Means t-SNE & unweighted cluster mean & \(0.708 \pm 0.189\) & \(1.095 \pm 0.572\) & \(0.307 \pm 0.092\) \\
        \bottomrule
    \end{tabular}
\end{table*}

\section{Discussion}

In this study, we applied unsupervised clustering to a large, unlabeled dataset to extract meaningful insights without prior knowledge of the data. In real-world applications, experts often lack the time and resources to manually review each individual sample or signal. The goal of our approach is to provide a general understanding of the dataset and identify a relevant subset for further investigation. To the best of our knowledge, no similar methods have been applied in this context, and we propose a systematic approach to address this gap.

Our methodology involved using the FFT to extract frequency-domain features, followed by PCA and t-SNE to obtain reduced representations for clustering. K-Means was then applied to both representations.

From the initial clustering results, two key observations emerged. First, clustering directly on PCA features resulted in highly imbalanced cluster sizes (see Table~\ref{tab:number_samples_per_cluster_pca}), while applying K-Means to the t-SNE features—after additional reduction via PCA—yielded more balanced cluster sizes (see Table~\ref{tab:number_samples_per_cluster_tsne}).

For a quantitative evaluation, we computed several clustering performance metrics, including purity, entropy, and silhouette score. The purity scores were relatively high, with values of \(0.608\) for PCA and \(0.721\) for t-SNE, indicating that the clusters are somewhat homogeneous with respect to the fault types. The silhouette scores, computed in the same 3-component representation in which each clustering was obtained, are \(0.487\) for PCA and \(0.315\) for t-SNE. Both are positive, indicating that the partitions are geometrically coherent in their own feature spaces, with the PCA partition the more compact of the two. Computed over the full dataset rather than the reviewed subset the values are \(0.501\) and \(0.326\), so this is not an artifact of the subset. Geometric coherence should not be read as agreement with the expert categories: the purity and entropy figures above show that the clusters remain internally mixed with respect to event type. Purity and entropy are computed on the reviewed subset (\(N=204\)) and may not fully reflect performance on the entire dataset.

When analyzing the individual clusters, it is essential to consider the varying cluster sizes. For PCA-based clusters, the largest cluster, containing 80 reviewed samples, has a purity of only \(0.425\), yet the highest per-cluster silhouette of the PCA partition, \(0.729\). It is thus a compact, well-separated region of the feature space that nonetheless spans several expert categories, which illustrates directly that geometric coherence and agreement with the expert taxonomy are distinct properties. In contrast, for t-SNE-based clusters, the largest cluster (cluster 0) achieves a much higher purity of \(0.864\) with a silhouette of \(0.385\), despite containing only 22 reviewed samples.

The summary statistics for K-Means clustering using both PCA and t-SNE features are provided in Table~\ref{tab:clustering_metrics}. Two aggregation conventions are reported separately and are not mixed within a row. Under the unweighted cluster mean, K-Means with PCA features gives a purity of \(0.730 \pm 0.216\), an entropy of \(0.767 \pm 0.585\), and a silhouette of \(0.371 \pm 0.201\); with t-SNE features, a purity of \(0.708 \pm 0.189\), an entropy of \(1.095 \pm 0.572\), and a silhouette of \(0.307 \pm 0.092\).

Under size weighting, K-Means with PCA features gives a purity of \(0.608 \pm 0.182\), an entropy of \(1.336 \pm 0.555\) and a silhouette of \(0.487 \pm 0.212\); with t-SNE features, a purity of \(0.721 \pm 0.184\), an entropy of \(1.073 \pm 0.557\) and a silhouette of \(0.315 \pm 0.085\). The size-weighted figures are the ones quoted in Table~\ref{tab:global_clustering_metrics}.

The two representations therefore rank differently depending on the criterion. The t-SNE-based clusters are more balanced in size and agree better with the expert event types (purity \(0.721\) against \(0.608\)), whereas the PCA-based partition is the more strongly separated of the two in its own feature space (silhouette \(0.487\) against \(0.315\)). Taken together, the unsupervised representations exhibit meaningful but incomplete cluster structure: the waveform clusters capture real structure in the signals, and that structure is not identical to the expert taxonomy.

Figures~\ref{fig:event_type_vs_kmeans_pca_heatmap_percentage} and~\ref{fig:event_type_vs_kmeans_tsne_heatmap_percentage} show that some clusters (in both methods) consist predominantly of a single fault type, such as short-circuit or switching events. The most notable difference between the two feature reduction methods is that the largest PCA cluster, which contains 80 samples from all fault types, fails to capture the distinct fault categories. In contrast, K-Means clustering based on the t-SNE features better distinguishes between different fault types. For further reference, the absolute numbers are provided in Appendix Figures~\ref{fig:event_type_vs_kmeans_pca_heatmap} and~\ref{fig:event_type_vs_kmeans_tsne_heatmap}.

\section{Conclusion}

This study demonstrates that K-Means clustering can effectively identify fault patterns in voltage and current signals, providing a structured approach to analyzing large-scale, unlabeled datasets. Expert validation of a subset of \numprint{204} samples out of \numprint{12053} confirmed that the identified clusters partially align with known fault types. These results highlight the potential of unsupervised learning for fault analysis in power systems.

A key advantage of this approach is its ability to segment large datasets into smaller, structured subsets, facilitating targeted analysis. Despite its limitations, clustering enables engineers to efficiently filter signals of interest. Unsupervised clustering reduces the number of samples requiring detailed inspection, thereby improving diagnostic efficiency and enabling a more systematic evaluation of fault characteristics.

However, several limitations must be considered. The absence of prior knowledge about the dataset limits the interpretability of the clustering results, as the extracted features may not fully capture the underlying fault characteristics. Additionally, the validation process is restricted by the labor-intensive nature of manual labeling, with only \numprint{204} of the \numprint{12053} recordings (\(1.7\%\)) being inspected. While these results offer an initial assessment, their statistical power remains limited, highlighting the need for larger labeled datasets for comprehensive evaluation.

Methodologically, K-Means assumes spherical cluster structures and equal cluster sizes, which may not align with the true distribution of fault patterns. The positive silhouette scores indicate that the partitions are geometrically coherent within their reduced representations, but the purity and entropy figures show only partial alignment with the expert event categories. The limiting factor therefore appears to be the representation and the assumption that fault characteristics dominate it, rather than cluster compactness alone. Density-based or spectral methods, and representations learned with fault structure in mind, remain worth investigating. 

A key challenge in this study is the assumption that fault characteristics dominate the clustering process. In reality, hidden variables such as variations in recording devices, sensor locations, or environmental conditions may introduce significant, unmeasured influences. Without explicit control over these factors, their impact on clustering performance remains unknown, potentially introducing bias and reducing interpretability.

Our contribution lies in proposing a systematic approach for unsupervised clustering of voltage and current signals to identify fault patterns without prior knowledge of the dataset. By applying K-Means clustering to a large, unlabeled dataset and validating the results with expert feedback, we demonstrate the feasibility of unsupervised learning in fault analysis. This approach provides a scalable and structured method for analyzing large datasets, which can be further enhanced by integrating simulated fault scenarios for controlled validation.

Future research should explore hybrid models that integrate unsupervised and supervised learning to improve clustering accuracy. Incorporating simulated fault scenarios could refine the clustering process by providing controlled conditions for validation.

While unsupervised clustering cannot replace expert-driven fault classification, it offers a systematic, scalable approach for analyzing large datasets, contributing to more reliable data-driven approaches for fault analysis in high-voltage power systems. Addressing the identified challenges will further enhance the robustness of clustering techniques in this context.

\section*{Acknowledgment}
This project was funded by the Deutsche Forschungsgemeinschaft (DFG, German Research Foundation) - 535389056.

\bibliographystyle{IEEEtran}
\bibliography{refs}

\vfill \pagebreak

% Author Biographies
\section*{Biographies}
\noindent
\small \textbf{Julian Oelhaf} is a PhD student at the Pattern Recognition Lab, Friedrich-Alexander-Universität Erlangen-Nürnberg (FAU), under the supervision of Dr.-Ing. Siming Bayer and Prof. Dr.-Ing. habil. Andreas Maier. He began his doctoral studies in 2024 after completing a Master’s degree in Computer Science at FAU. His research focuses on machine learning applications for fault detection and protection coordination.
\\[0.5cm] % Adds space after each biography

\noindent
\textbf{Georg Kordowich} studied Electrical Energy Technologies at the Friedrich-Alexander-University Erlangen-Nürnberg. After completing his Master’s Degree in 2021 he started working as a PhD Student at the Institute of Electrical Energy Systems at FAU under the supervision of Prof. Dr. Johann Jäger. 
His main research focus is the application of machine learning for fault detection and protection coordination. \\[0.5cm] % Adds space after each biography

\noindent
\textbf{Andreas Maier} (M’05–SM’20) received a M.Sc. in computer science from the University of Erlangen–Nuremberg, Erlangen, Germany, in 2005, and the Ph.D. degree in computer science from the same institution in 2009. He is currently a Professor and Head of the Pattern Recognition Lab at the University of Erlangen–Nuremberg, where he has been a faculty member since 2015. His research interests include medical imaging, image and audio processing, digital humanities, and interpretable machine learning. Prof. Maier has developed significant tools in medical signal processing, such as the PEAKS online speech intelligibility assessment tool, and has contributed to over 4,000 patient and control subject analyses. He has received numerous awards, including an ERC Synergy Grant for “4D Nanoscope”.\\[0.5cm] % Adds space after each biography

\noindent
\textbf{Johann Jäger} received the Dipl.-Ing. and Dr.-Ing. degrees in electrical engineering from Friedrich-Alexander-University Erlangen-
Nuremberg (FAU), Erlangen, Germany, in 1990 and 1996, respectively. From 1996, he was with Power System Planning Department, Siemens, Erlangen, Germany. He was working on different fields of FACTS devices, network planning and protection systems in worldwide projects. Since 2004, he is in charge of a Full Professorship for Electrical Power Systems with the FAU University Erlangen-Nürnberg. He is author of numerous publications and books as well as inventor of several patents.\\[0.5cm] % Adds space after each biography

\noindent  
\textbf{Siming Bayer} received her Dr.-Ing. in Computer Science from Friedrich-Alexander-Universität Erlangen-Nürnberg (FAU) in 2022, where she developed machine learning algorithms for vascular structure registration in collaboration with Siemens Healthineers. She also holds an M.Sc. in Biomedical Engineering from FAU. 
She is currently a Research Scientist at FAU, leading the ''Data Processing for Utility Infrastructure'' group at the Pattern Recognition Lab, and a Strategic Research \& Collaboration Manager at Siemens Healthineers, focusing on cross-business research in digital health. Her research interests include machine learning for medical imaging, industrial data processing, and innovation management in healthcare and utility infrastructures.
\\[0.5cm] % Adds space after each biography

\vfill \pagebreak
\appendix

\begin{figure}[!hbp]
    \centering
    \includegraphics[width=0.6\linewidth]{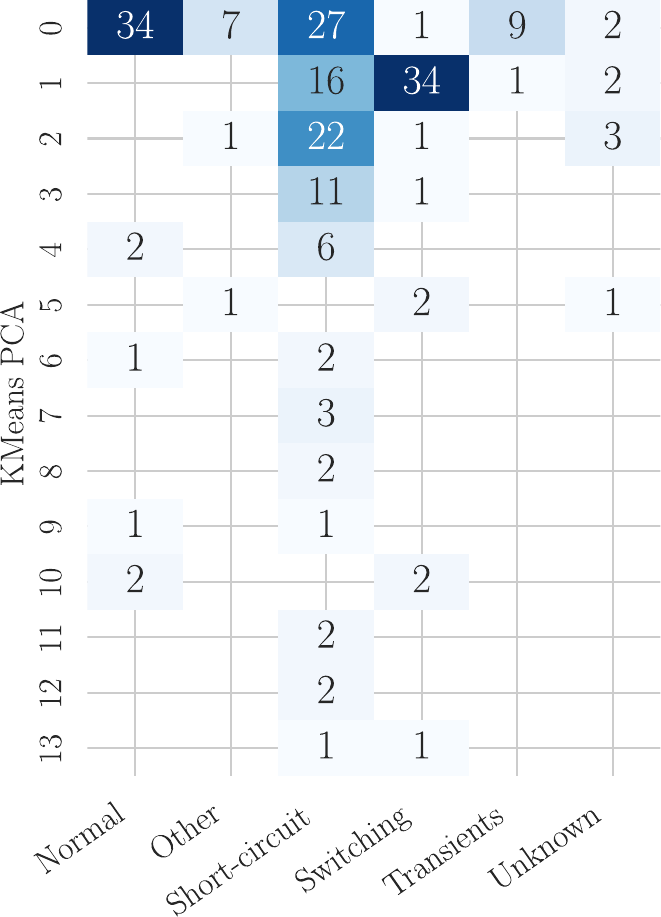}
    \caption{Absolute number of event types per K-Means PCA cluster in the manually labeled subset (\(N = 204\))}
    \label{fig:event_type_vs_kmeans_pca_heatmap}
\end{figure}

\begin{figure}[!hbp]
    \centering
    \includegraphics[width=0.6\linewidth]{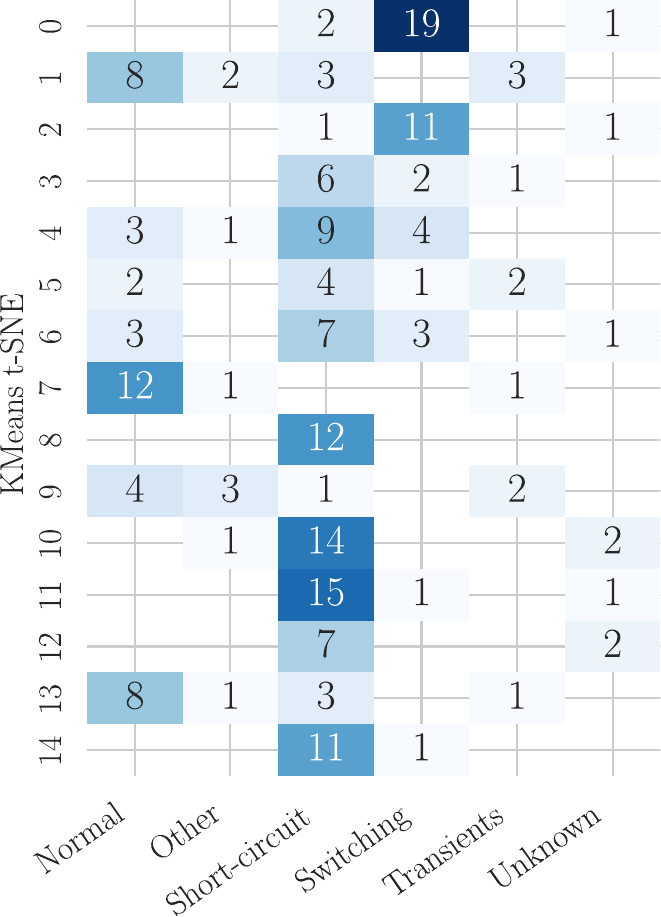}
    \caption{Absolute number of event types per K-Means t-SNE cluster in the manually labeled subset (\(N = 204\))}
    \label{fig:event_type_vs_kmeans_tsne_heatmap}
\end{figure}

\begin{figure*}[!hb]
	\centering
	\includegraphics[trim={0 0 0 0},clip,width=0.9\linewidth]{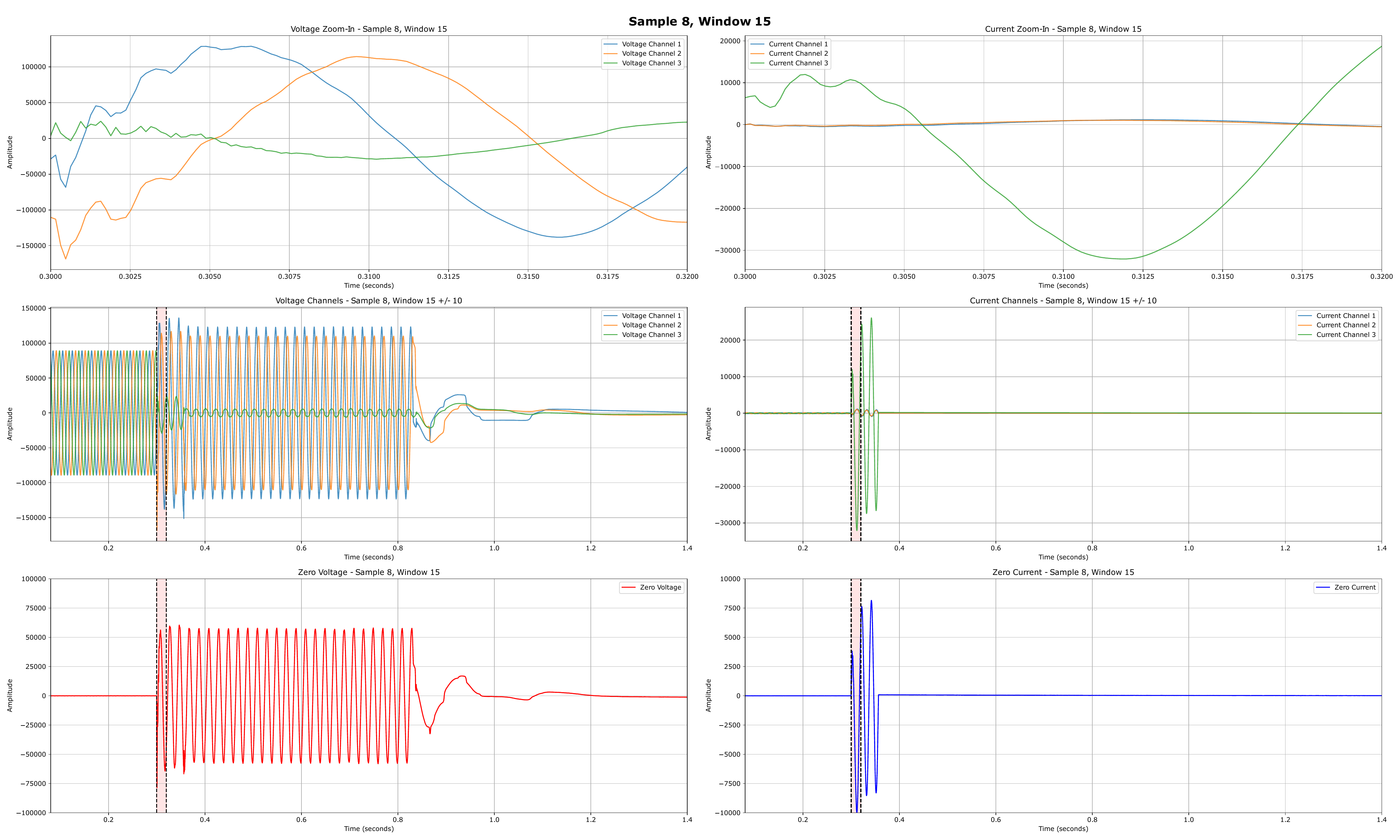}
    \caption{Visualization of voltage and current signals for Sample 8, Window 15. The figure presents a detailed time-series analysis of three-phase voltage and current waveforms, capturing fault characteristics in a high-voltage power system. The top row provides zoomed-in views of the voltage (left) and current (right) signals within the fault event window. The middle row shows a broader perspective of the voltage (left) and current (right) waveforms surrounding the fault event, marked by vertical dashed lines. The bottom row highlights the zero-sequence voltage (left) and current (right), which help in detecting fault signatures. These signals are used to cluster and analyze fault events in an unsupervised manner.}
    \label{fig:voltage_current_analysis}
\end{figure*}

% Generated by fault_cluster.reports.fda2025 -- do not edit by hand.
% Canonical run: 2026-09-09_fda2025_v2_8a2c1d279141
% Silhouette is computed in the representation each clustering was fitted in.
\begin{table*}[!htbp]
\caption{Per-Cluster Clustering Evaluation Metrics (\(N=204\))}
\label{tab:per_cluster_clustering_metrics}
\centering
\begin{tabular}{lrrrrr}
\toprule
\textbf{Clustering Method} & \textbf{Cluster} & \textbf{Count} & \makecell{\textbf{Purity} ($\uparrow$)} & \makecell{\textbf{Entropy} ($\downarrow$)} & \makecell{\textbf{Silhouette} ($\uparrow$)} \\
\midrule
\multirow{14}{*}{K-Means PCA}
 & 0  & 80  & 0.425 & 1.928 & 0.729 \\
 & 1  & 53  & 0.642 & 1.219 & 0.304 \\
 & 2  & 27  & 0.815 & 0.945 & 0.405 \\
 & 3  & 12  & 0.917 & 0.414 & 0.248 \\
 & 4  & 8   & 0.750 & 0.811 & 0.179 \\
 & 5  & 4   & 0.500 & 1.500 & 0.642 \\
 & 6  & 3   & 0.667 & 0.918 & 0.269 \\
 & 7  & 3   & 1.000 & 0.000 & 0.405 \\
 & 8  & 2   & 1.000 & 0.000 & 0.093 \\
 & 9  & 2   & 0.500 & 1.000 & 0.088 \\
 & 10 & 4   & 0.500 & 1.000 & 0.501 \\
 & 11 & 2   & 1.000 & 0.000 & 0.180 \\
 & 12 & 2   & 1.000 & 0.000 & 0.552 \\
 & 13 & 2   & 0.500 & 1.000 & 0.596 \\
\midrule
\multirow{15}{*}{K-Means t-SNE}
 & 0  & 22  & 0.864 & 0.700 & 0.385 \\
 & 1  & 16  & 0.500 & 1.781 & 0.317 \\
 & 2  & 13  & 0.846 & 0.773 & 0.244 \\
 & 3  & 9   & 0.667 & 1.224 & 0.140 \\
 & 4  & 17  & 0.529 & 1.659 & 0.297 \\
 & 5  & 9   & 0.444 & 1.837 & 0.322 \\
 & 6  & 14  & 0.500 & 1.724 & 0.321 \\
 & 7  & 14  & 0.857 & 0.735 & 0.473 \\
 & 8  & 12  & 1.000 & 0.000 & 0.262 \\
 & 9  & 10  & 0.400 & 1.846 & 0.459 \\
 & 10 & 17  & 0.824 & 0.834 & 0.343 \\
 & 11 & 17  & 0.882 & 0.640 & 0.315 \\
 & 12 & 9   & 0.778 & 0.764 & 0.208 \\
 & 13 & 13  & 0.615 & 1.489 & 0.164 \\
 & 14 & 12  & 0.917 & 0.414 & 0.361 \\
\bottomrule
\end{tabular}
\end{table*}

\end{document}